# Iterative Join-Graph Propagation

Rina Dechter    Kalev Kask    Robert Mateescu
Department of Information and Computer Science
University of California, Irvine, CA 92697-3425
{dechter,kkask,mateescu}@ics.uci.edu

## Abstract

The paper presents an iterative version of join-tree clustering that applies the message passing of join-tree clustering algorithm to join-graphs rather than to join-trees, iteratively. It is inspired by the success of Pearl's belief propagation algorithm (BP) as an iterative approximation scheme on one hand, and by a recently introduced mini-clustering (MC(i)) success as an anytime approximation method, on the other. The proposed *Iterative Join-graph Propagation (IJGP)* belongs to the class of generalized belief propagation methods, recently proposed using analogy with algorithms in statistical physics. Empirical evaluation of this approach on a number of problem classes demonstrates that even the most time-efficient variant is almost always superior to IBP and MC(i), and is sometimes more accurate by as much as several orders of magnitude.

## 1 INTRODUCTION

Probabilistic reasoning using Belief networks, computing the probability of one or more events given some evidence, is known to be NP-hard [Cooper1990]. However most commonly used exact algorithms for probabilistic inference such as join-tree clustering [Lauritzen and Spiegelhalter1988, Jensen et al.1990] or variable-elimination [Dechter1996] exploit the networks structure. These algorithms are time and space exponential in a graph parameter capturing the density of the network called tree-width. Yet, for large belief networks, the tree-width is often large, making exact inference impractical and therefore approximation methods must be pursued. Although approximation within given error bounds is also NP-hard [Dagum and Luby1993, Roth1996], some approximation strategies work well in practice. One promising methodology pursued in such hard computational cases is that of developing anytime algorithms.

Belief propagation (BP) algorithm is a distributed algorithm that computes posterior beliefs for tree-structured Bayesian networks. However, in recent years it was shown to work surprisingly well in many applications involving networks with loops, including turbo codes, when applied iteratively. While there is still very little understanding as to why and when Iterative Belief Propagation (IBP) will work well, some progress was made recently in understanding the algorithm's behavior, showing that it converges to a stationary point of Bethe energy, thus making connection to approximation algorithms developed in statistical physics and to variational approaches to approximate inference [Welling and Teh2001, Yedidia et al.2001].

The main shortcoming of IBP however, is that it cannot be improved if allowed more time. Namely, it does not have an anytime behavior. Approximation algorithms further developed in statistical physics called Kikuchi approximations improve over Bethe energy showing how to construct more accurate free energy approximations. Exploiting this insight, [Yedidia et al.2001] proposed Generalized Belief Propagation (GBP) approximations that extend IBP towards being an anytime algorithm, and provided some initial empirical demonstration that these new algorithms can be significantly more accurate than ordinary IBP at an adjustable increased complexity. The central idea is to improve approximation by clustering some of the network's nodes into super nodes and apply message passing between the super nodes rather than between the original singleton nodes.

We will present in this paper a special class of GBP algorithms called *Iterative Join-Graph Propagation (IJGP(i))* which are controlled by a bounding parameter $i$ that allows the user to control the tradeoff between time and accuracy. The algorithm exploits an intuition based solely on concepts and algorithms developed within the theory and practice of belief net-works, and thus may be more accessible. The idea is as follows.

Pearl's BP algorithm on trees was extended to a general propagation algorithm on trees of clusters called join-tree clustering or junction-tree clustering [Lauritzen and Spiegelhalter1988, Jensen et al.1990]. Since this join-tree clustering is a message passing algorithm between clusters of functions, it can also be applied to a join-graph rather than a join-tree. Namely, rather than decomposing the network into a join-tree whose clusters are often too big and thus too costly to process, we can decompose the network into a join-graph having manageable clusters and apply join-tree message-passing over the join-graph, iteratively.

The question we explore is how will IJGP(i) work on join-graphs having cluster size bounded by $i$ variables and to what extent the algorithm is sensitive to the particular join-graph selected. We hypothesize that as the decomposition is more coarse we get more accurate performance, yielding an anytime behavior at an adjusted increased complexity.

Algorithm IJGP(i) can also be seen as an iterative version of mini-clustering MC(i), a recently proposed anytime approximation for belief updating [Dechter and Rish1997, Mateescu et al.2002], which was shown to be competitive with IBP and Gibbs sampling on a variety of benchmarks. Mini-clustering algorithm partitions the messages passed in the join-tree between clusters. Namely, instead of computing and sending one message over the separator between two clusters, MC(i) sends a set of smaller messages, each computed by a mini-partition in the cluster, and each defined on no more than $i$ variables.

When we started experimenting with IJGP it became clear immediately that using arc-minimal join-graph is essential to the success of IJGP(i). However it appears that arc-minimality is not always sufficient, yielding a refined definition of arc-labeled join-graph capturing the need to avoid cyclicity relative to every single variable.

Following preliminaries, we give a formal account of arc-labeled parameterized join-graph decomposition and define IJGP(i) over such decompositions. Some properties of the algorithm are discussed. Subsequently we provide empirical evaluation. The empirical results are very encouraging. We demonstrate that even for i=2, when IJGP is fastest, the algorithm is already very effective. Overall, it is an anytime scheme which outperforms IBP and MC(i), sometimes by as much as several orders of magnitude.

## 2 PRELIMINARIES

*Belief networks* provide a formalism for reasoning about partial beliefs under conditions of uncertainty. A belief network is defined by a directed acyclic graph over nodes representing random variables. The family of $X_i$, $F_i$, includes $X_i$ and its parent variables.
**Belief networks.** A *belief network* is a quadruple $BN = <X, D, G, P>$ (also abbreviated $<G, P>$ when $X$ and $D$ are clear) where $X = \{X_1, \ldots, X_n\}$ is a set of random variables, $D = \{D_1, \ldots, D_n\}$ is the set of the corresponding domains, $G$ is a directed acyclic graph over $X$ and $P = \{p_1, \ldots, p_n\}$, where $p_i = P(X_i|pa_i)$ ($pa_i$ are the parents of $X_i$ in $G$) denote conditional probability tables (CPTs). Given a function $f$, we denote by $scope(f)$ the set of its arguments.
**Belief updating.** The *belief updating* problem defined over a belief network (also referred to as *probabilistic inference*) is the task of computing the posterior probability $P(Y|e)$ of *query* nodes $Y \subseteq X$ given evidence $e$. We will focus on the basic case when $Y$ consists of a single variable $X_i$. Namely, on computing $Bel(X_i) = P(X_i = x|e), \forall X_i \in X, \forall x \in D_i$.

## 3 JOIN-GRAPHS

We will describe our algorithms relative to a join-graph decomposition framework using recent notation proposed by [Gottlob et al.1999]. The notion of join-tree decompositions was introduced in relational databases [Maier1983].

DEFINITION **3.1 (join-graph decompositions)** *A join-graph decomposition for $BN = <X, D, G, P>$ is a triple $D = <JG, \chi, \psi>$, where $JG = (V, E)$ is a graph, and $\chi$ and $\psi$ are labeling functions which associate with each vertex $v \in V$ two sets, $\chi(v) \subseteq X$ and $\psi(v) \subseteq P$ such that,*
*1. For each function $p_i \in P$, there is exactly one vertex $v \in V$ such that $p_i \in \psi(v)$, and $scope(p_i) \subseteq \chi(v)$.*
*2. (connectedness) For each variable $X_i \in X$, the set $\{v \in V | X_i \in \chi(v)\}$ induces a connected subgraph of $G$. The connectedness requirement is also called the running intersection property.*

We will often refer to a node and its CPT functions as a *cluster*[1] and use the term *join-graph-decomposition* and *cluster graph* interchangeably. A *join-tree-decomposition* or a *cluster tree* is the special case when the join-graph JG is a tree.

**Join-tree propagation**. The well known join-tree clustering algorithm first converts the belief network into a cluster tree and then sends messages between clusters. We call the second message passing phase "join-tree propagation". The complexity of join-tree clustering is exponential in the number of variables in a cluster (tree-width), and the number of variables in the intersections between adjacent clusters (separator-width), as defined below.

---
[1]Note, that a node may be associated with an empty set of CPTs

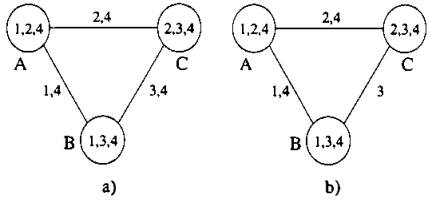

Figure 1: An arc-labeled decomposition

DEFINITION 3.2 (tree-width, separator-width)
Let $D = <JT, \chi, \psi>$ be a tree decomposition of a belief network $<G, P>$. The tree-width of $D$ [Arnborg1985] is $max_{v \in V} |\chi(v)|$. The tree-width of $<G, P>$ is the minimum tree-width over all its join-tree decompositions. Given two adjacent vertices $u$ and $v$ of $JT$, the separator of $u$ and $v$ is defined as $sep(u, v) = \chi(u) \cap \chi(v)$, and the separator-width is $max_{(u,v)} |sep(u, v)|$.

The minimum tree-width of a graph $G$ can be shown to be identical to a related parameter called induced-width. A join-graph decomposition $D$ is *arc-minimal* if none of its arcs can be removed while still satisfying the connectedness property of Definition 3.1. In our preliminary experiments we observed immediately that when applying tree propagation on join-graph iteratively, it is crucial to avoid cycling messages relative to every single variable. The property of arc-minimality is *not* sufficient to ensure such acyclicity though.

**Example 3.1** *The example in Figure 1a shows an arc minimal join-graph which contains a cycle relative to variable 4, with arcs labeled with separators. Notice however that if we remove variable 4 from the label of one arc we will have no cycles (relative to single variables) while the connectedness property will still be maintained.*

To allow more flexible notions of connectedness we refine the definition of join-graph decompositions, when arcs can be labeled with a subset of their separator.

DEFINITION 3.3 (arc-labeled join-graph decompositions) An arc-labeled decomposition *for* $BN = <X, D, G, P>$ *is a four-tuple* $D = <JG, \chi, \psi, \theta>$, *where* $JG = (V, E)$ *is a graph*, $\chi$ *and* $\psi$ *associate with each vertex* $v \in V$ *the sets* $\chi(v) \subseteq X$ *and* $\psi(v) \subseteq P$ *and* $\theta$ *associates with each edge* $(v, u) \subset E$ *the set* $\theta((v, u)) \subseteq X$ *such that:*
1. *For each function* $p_i \in P$, *there is* exactly *one vertex* $v \in V$ *such that* $p_i \in \psi(v)$, *and* $scope(p_i) \subseteq \chi(v)$.
2. *(arc-connectedness) For each arc* $(u, v)$, $\theta(u, v) \subseteq sep(u, v)$, *such that* $\forall X_i \in X$, *any two clusters containing* $X_i$ *can be connected by a path whose every arc's label includes* $X_i$.

*Finally, an arc-labeled join graph is* minimal *if no variable can be deleted from any label while still satisfying the arc-connectedness property.*

DEFINITION 3.4 (eliminator) *Given two adjacent vertices* $u$ *and* $v$ *of* $JG$, *the* eliminator *of* $u$ *with respect to* $v$ *is* $elim(u, v) = \chi(u) - \theta((u, v))$.

Arc-labeled join-graphs can be made minimal by deleting variables from the labels. It is easy to see that a *minimal arc-labeled join-graph* does not contain any cycle relative to any single variable. That is, any two clusters containing the same variable are connected by exactly one path labeled with that variable.

## 4 ALGORITHM ITERATIVE JOIN-GRAPH PROPAGATION

Applying join-tree propagation iteratively to join-graphs yields algorithm *Iterative Join-Graph Propagation (IJGP)* described in Figure 2. One iteration of the algorithm applies message-passing in a topological order over the join-graph, forward and back.

When node $i$ sends a message (or messages) to a neighbor node $j$ it operates on all the CPTs in its cluster and on all the messages sent from its neighbors excluding the ones received from $j$. First, all individual functions that share no variables with the eliminator are collected and sent to $j$. All the rest of the functions are *combined* in a product and summed over the eliminator between $i$ and $j$. Figures 3 and 4 describe a belief network, a join-tree decomposition and the trace of running IJGP over a join-tree. Indeed, it is known that:

THEOREM 4.1 *1. [Lauritzen and Spiegelhalter1988] If IJGP is applied to a join-tree decomposition it reduces to join-tree clustering and it therefore is guaranteed to compute the exact beliefs in one iteration.*
*2. [Larrosa et al.2001] The time complexity of one iteration of IJGP is $O(deg \cdot (n + N) \cdot d^{w^*+1})$ and its space complexity is $O(N \cdot d^\theta)$, where deg is the maximum degree of a node in the join-graph, n is the number of variables, N is the number of nodes in the graph decomposition, d is the maximum domain size, $w^*$ is the maximum cluster size and $\theta$ is the maximum label size.*

However, when applied to a join-graph the algorithm is neither guaranteed to converge nor to find the exact posterior.

The success of IJGP, no doubt, will depend on the

**Algorithm IJGP**
**Input:** An arc-labeled join-graph decomposition $< JG, \chi, \psi, \theta >, JG = (V, E)$ for $BN =< X, D, G, P >$. Evidence variables $var(e)$.
**Output:** An augmented graph whose nodes are clusters containing the original CPTs and the messages received from neighbors. Approximations of $P(X_i|e), \forall X_i \in X$.

Denote by $h_{(u,v)}$ the message from vertex $u$ to $v$, $ne_v(u)$ the neighbors of $u$ in $JG$ excluding $v$.
$cluster(u) = \psi(u) \cup \{h_{(v,u)}|(v,u) \in E\}$.
$cluster_v(u) = cluster(u)$ excluding message from $v$ to $u$.
• **One iteration of IJGP**
**For** every node $u$ in $JG$ in some topological order $d$ and back, do
1. **process observed variables**
Assign relevant evidence to all $p_i \in \psi(u)$ $\chi(u) := \chi(u) - var(e), \forall u \in V$
2. **Compute individual functions:** Include in $H_{(u,v)}$ each function in $cluster_v(u)$ whose scope does not contain variables in $elim(u,v)$. Denote by $A$ the remaining functions.
3. **Compute and send to $v$ the combined function:** $h_{(u,v)} = \sum_{elim(u,v)} \prod_{f \in A} f$.
Send $h_{(u,v)}$ and the individual functions $H_{(u,v)}$ to node $v$.
**Endfor**
• **Compute $P(X_i, e)$:**
For every $X_i \in X$ let $u$ be a vertex in $T$ such that $X_i \in \chi(u)$. Compute $P(X_i|e) = \alpha \sum_{\chi(u)-\{X_i\}} (\prod_{f \in cluster(u)} f)$

Figure 2: Algorithm iterative join-graph propagation

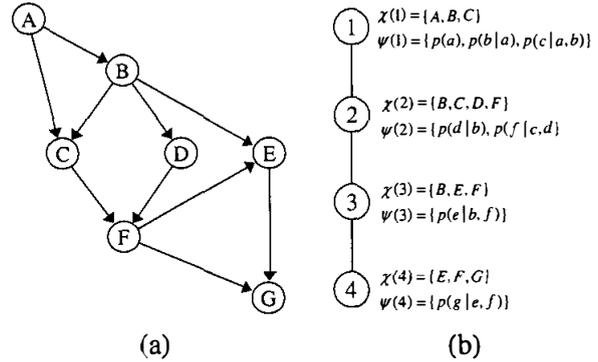

(a) 　　　　　　　　　　(b)

Figure 3: a) A belief network; b) A join-tree decomposition

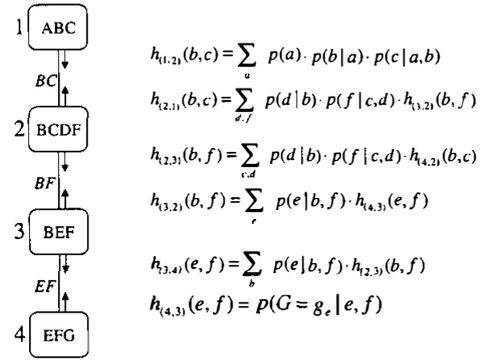

Figure 4: Execution of IJGP on a join-tree

choice of cluster graphs it operates on. The following paragraphs provide some rationale to our choice of minimal arc-labeled join-graphs. First, we are committed to the use of an underlying graph structure that capture as many of the distribution independence relations as possible, without introducing new ones. That is, we restrict attention to cluster graphs that are I-maps of $P$ [Pearl1988]. Second, we wish to avoid cycles as much as possible in order to minimize computational overcounting.

Indeed, it can be shown that any join-graph of a belief network is an I-map of the underlying probability distribution relative to node-separation. It turns out that arc-labeled join-graphs display a richer set of independencies relative to arc-separation.

DEFINITION **4.1 (arc-separation in (arc-labeled) join-graphs)** Let $D =< JG, \chi, \psi, \theta >$, $JG = (V, E)$ be an arc-labeled decomposition. Let $N_W, N_Y \subseteq V$ be two sets of nodes, and $E_Z \subseteq E$ be a set of edges in $JG$. Let $W, Y, Z$ be their corresponding sets of variables $(W = \cup_{v \in N_W} \chi(v), Z = \cup_{e \in E_Z} \theta(e))$. $E_Z$ arc-separates $N_W$ and $N_Y$ in $D$ if there is no path between $N_W$ and $N_Y$ in the graph $JG$ with the edges in $E_Z$ removed. In this case we also say that $W$ is separated from $Y$ given $Z$ in $D$. Arc-separation in a regular join-graph is defined relative to its separators.

Interestingly however, removing arcs or labels from arc-labeled join-graphs whose clusters are fixed will not increase the independencies captured by arc-labeled join-graphs. That is, any two (arc-labeled) join-graphs defined on the same set of clusters, sharing $(V, \chi, \psi)$, express exactly the same set of independencies relative to arc-separation.

THEOREM **4.2** *Any arc-labeled join graph decomposition of a belief network $BN =< X, D, G, P >$ is a minimal I-map of $P$ relative to arc-separation.*

Hence, the issue of minimizing computational overcounting due to cycles appears to be orthogonal to maximizing independencies via minimal I-mappness. Nevertheless, to avoid over-counting as much as possible, we still prefer join-graphs that minimize cycles relative to each variable. That is, we prefer to apply IJGP to *minimal* arc-labeled join-graphs.

## 5　BOUNDED JOIN-GRAPHS

Since we want to control the complexity of IJGP we will define it on decompositions having bounded cluster size. If the number of variables in a cluster is bounded by $i$, the time and space complexity of one full iteration of IJGP(i) is exponential in $i$. How can good graph-decompositions of bounded cluster size be generated?

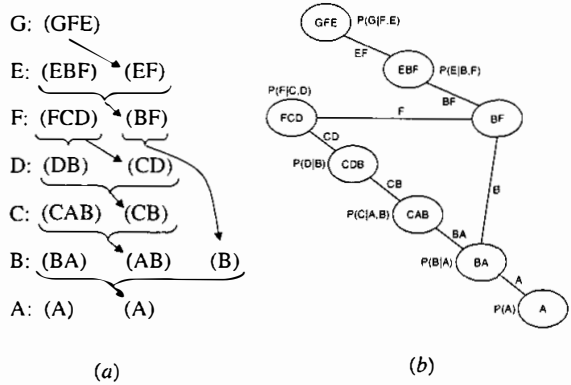

Figure 5: Join-graph decompositions

---

**Algorithm join-graph structuring(i)**

1. Apply procedure schematic mini-bucket(i).
2. Associate each resulting mini-bucket with a node in the join-graph, the variables of the nodes are those appearing in the mini-bucket, the original functions are those in the mini-bucket.
3. Keep the arcs created by the procedure (called out-edges) and label them by the regular separator.
4. Connect the mini-bucket clusters belonging to the same bucket in a chain by in-edges labeled by the single variable of the bucket.

---

Figure 6: Algorithm join-graph structuring(i)

Since we want the join-graph to be as close as possible to a tree, and since a tree has a tree-width 1, we may try to find join-graph $JG$, of bounded cluster size whose tree-width (as a graph) is minimized. While we will not attempt to optimally solve this task, we will propose one method for generating i-bounded graph-decomposition.

One class of such decompositions is partition-based. It starts from a given tree-decomposition and then partitions the clusters until the decomposition has clusters bounded by $i$. The opposite approach is grouping-based. It starts from an arc-minimal dual-graph decomposition (where each cluster contains a single CPT) and groups clusters into larger clusters as long as the resulting clusters do not exceed the given bound. In both methods we should attempt to reduce the tree-width of the generated graph-decomposition. Our partition-based approach inspired by the mini-bucket idea [Dechter and Rish1997] is as follows.

Given a bound $i$, algorithm *join-graph structuring(i)* applies procedure *schematic mini-bucket(i)*, described in Figure 7. The procedure only traces the scopes of the functions that would be generated by the full mini-bucket procedure, avoiding actual computation. The algorithm then connects the mini-buckets' scopes minimally to obtain the running intersection property, as

---

**Procedure schematic mini-bucket(i)**

1. Order the variables from $X_1$ to $X_n$ minimizing (heuristically) induced-width, and associate a bucket for each variable.
2. Place each CPT in the bucket of the highest index variable in its scope.
3. For $j = n$ to 1 do:
   Partition the functions in $bucket(X_j)$ into mini-buckets having at most $i$ variables.
   For each mini-bucket $mb$ create a new scope-function (message) $f$ where $scope(f) = \{X | X \in mb\} - \{X_i\}$ and place scope(f) in the bucket of its highest variable. Maintain an arc between $mb$ and the mini-bucket (created later) of $f$.

---

Figure 7: Procedure schematic mini-bucket(i)

described in Figure 6.

**Example 5.1** *Figure 5a shows the trace of procedure schematic mini-bucket(3) applied to the problem described in Figure 3. The decomposition in Figure 5b is created by the algorithm graph structuring. The only cluster partitioned is that of F into two scopes (FCD) and (BF), connected by an in-edge labeled with F.*

Procedure schematic mini-bucket ends with a collection of trees rooted in mini-buckets of the first variable. Each of these trees is minimally arc-labeled. Then, *in-edges* are labeled with only one variable, and they are added only to obtain the running intersection property between branches of these trees. It can be shown that:

**Proposition 1** *Algorithm join-graph structuring(i), generates a minimal arc-labeled join-graph decomposition having bound $i$.*

**MC(i) vs. IJGP(i).** As can be hinted by our structuring of a bounded join-graph, there is a close relationship between MC(i) and IJGP(i). In particular, one iteration of IJGP(i) is similar to MC(i) (MC(i) is an algorithm that approximates join-tree clustering and was shown to be competitive with IBP and Gibbs Sampling [Mateescu et al.2002]). Indeed, while we view IJGP(i) as an iterative version of MC(i), the two algorithms differ in several technical points, some may be superficial, due to implementation, others may be more principled. We will leave the discussion at that and will observe the comparison of the two approaches in the empirical section.

## 6 EMPIRICAL EVALUATION

We tested the performance of IJGP(i) on random networks, on M-by-M grids, on two benchmark CPCS files with 54 and 360 variables, respectively (these are belief networks for medicine, derived from the Computer

Table 1: Random networks: N=50, K=2, C=45, P=3, 100 instances, w*=16

|     |       | Absolute error |     |     |     | Relative error |     |     |     | KL distance |     |     |     | Time |     |     |     |
|-----|-------|-----|-----|-----|-----|-----|-----|-----|-----|-----|-----|-----|-----|-----|-----|-----|-----|
|     |       | IBP | IJGP | | | IBP | IJGP | | | IBP | IJGP | | | IBP | IJGP | | |
| #it | #evid |     | i=2 | i=5 | i=8 |     | i=2 | i=5 | i=8 |     | i=2 | i=5 | i=8 |     | i=2 | i=5 | i=8 |
| 1   | 0     | 0.02988 | 0.03055 | 0.02623 | 0.02940 | 0.06388 | 0.15694 | 0.05677 | 0.07153 | 0.00213 | 0.00391 | 0.00208 | 0.00277 | 0.0017 | 0.0036 | 0.0058 | 0.0295 |
|     | 5     | 0.06178 | 0.04434 | 0.04201 | 0.04554 | 0.15005 | 0.12340 | 0.12056 | 0.11154 | 0.00812 | 0.00582 | 0.00478 | 0.00558 | 0.0013 | 0.0040 | 0.0052 | 0.0200 |
|     | 10    | 0.08762 | 0.05777 | 0.05409 | 0.05910 | 0.23777 | 0.18071 | 0.14278 | 0.15686 | 0.01547 | 0.00915 | 0.00768 | 0.00899 | 0.0013 | 0.0040 | 0.0036 | 0.0121 |
| 5   | 0     | 0.00829 | 0.00636 | 0.00592 | 0.00669 | 0.01726 | 0.01326 | 0.01219 | 0.01398 | 0.00021 | 0.00014 | 0.00015 | 0.00018 | 0.0066 | 0.0145 | 0.0226 | 0.1219 |
|     | 5     | 0.05182 | 0.00886 | 0.00886 | 0.01123 | 0.12589 | 0.01967 | 0.01965 | 0.02494 | 0.00658 | 0.00024 | 0.00026 | 0.00044 | 0.0060 | 0.0120 | 0.0185 | 0.0840 |
|     | 10    | 0.08039 | 0.01155 | 0.01073 | 0.01399 | 0.21781 | 0.03014 | 0.02553 | 0.03279 | 0.01382 | 0.00055 | 0.00042 | 0.00073 | 0.0048 | 0.0100 | 0.0138 | 0.0536 |
| 10  | 0     | 0.00828 | 0.00584 | 0.00514 | 0.00495 | 0.01725 | 0.01216 | 0.01069 | 0.01030 | 0.00021 | 0.00012 | 0.00010 | 0.00010 | 0.0130 | 0.0254 | 0.0436 | 0.2383 |
|     | 5     | 0.05182 | 0.00774 | 0.00732 | 0.00708 | 0.12590 | 0.01727 | 0.01628 | 0.01575 | 0.00658 | 0.00018 | 0.00017 | 0.00016 | 0.0121 | 0.0223 | 0.0355 | 0.1639 |
|     | 10    | 0.08040 | 0.00892 | 0.00808 | 0.00855 | 0.21782 | 0.02101 | 0.01907 | 0.02005 | 0.01382 | 0.00028 | 0.00024 | 0.00029 | 0.0109 | 0.0191 | 0.0271 | 0.1062 |
| MC  | 0     |     | 0.04044 | 0.04287 | 0.03748 |     | 0.08811 | 0.09342 | 0.08117 |     | 0.00403 | 0.00435 | 0.00369 |     | 0.0159 | 0.0173 | 0.0552 |
|     | 5     |     | 0.05303 | 0.05171 | 0.04250 |     | 0.12375 | 0.11775 | 0.09596 |     | 0.00659 | 0.00636 | 0.00477 |     | 0.0146 | 0.0158 | 0.0532 |
|     | 10    |     | 0.06033 | 0.05489 | 0.04266 |     | 0.14702 | 0.13219 | 0.10074 |     | 0.00841 | 0.00729 | 0.00503 |     | 0.0119 | 0.0143 | 0.0470 |

Table 2: CPCS networks: CPCS54 50 instances, w*=15; CPCS360 10 instances, w*=20

|     |       | Absolute error |     |     |     | Relative error |     |     |     | KL distance |     |     |     | Time |     |     |     |
|-----|-------|-----|-----|-----|-----|-----|-----|-----|-----|-----|-----|-----|-----|-----|-----|-----|-----|
|     |       | IBP | IJGP | | | IBP | IJGP | | | IBP | IJGP | | | IBP | IJGP | | |
| #it | #evid |     | i=2 | i=5 | i=8 |     | i=2 | i=5 | i=8 |     | i=2 | i=5 | i=8 |     | i=2 | i=5 | i=8 |
| CPCS54 | | | | | | | | | | | | | | | | | |
| 1   | 0     | 0.01324 | 0.03747 | 0.03183 | 0.02233 | 0.02716 | 0.08966 | 0.07761 | 0.05616 | 0.00041 | 0.00583 | 0.00512 | 0.00378 | 0.0097 | 0.0137 | 0.0146 | 0.0275 |
|     | 5     | 0.02684 | 0.03739 | 0.03124 | 0.02337 | 0.05736 | 0.09007 | 0.07676 | 0.05856 | 0.00199 | 0.00573 | 0.00493 | 0.00366 | 0.0072 | 0.0094 | 0.0087 | 0.0169 |
|     | 10    | 0.03915 | 0.03843 | 0.03426 | 0.02747 | 0.08475 | 0.09156 | 0.08246 | 0.06687 | 0.00357 | 0.00567 | 0.00506 | 0.00390 | 0.005  | 0.0047 | 0.0052 | 0.0115 |
| 5   | 0     | 0.00031 | 0.00016 | 0.00123 | 0.00110 | 0.00064 | 0.00033 | 0.00255 | 0.00225 | 7.75e-7 | 0.00000 | 0.00002 | 0.00001 | 0.0371 | 0.0334 | 0.0384 | 0.0912 |
|     | 5     | 0.01874 | 0.00058 | 0.00092 | 0.00098 | 0.04067 | 0.00124 | 0.00194 | 0.00203 | 0.00161 | 0.00000 | 0.00001 | 0.00001 | 0.0337 | 0.0215 | 0.0260 | 0.0631 |
|     | 10    | 0.03348 | 0.00101 | 0.00139 | 0.00144 | 0.07302 | 0.00215 | 0.00298 | 0.00302 | 0.00321 | 0.00001 | 0.00003 | 0.00002 | 0.0290 | 0.0144 | 0.0178 | 0.0378 |
| 10  | 0     | 0.00031 | 0.00009 | 0.00014 | 0.00015 | 0.00064 | 0.00018 | 0.00029 | 0.00031 | 7.75e-7 | 0.0000  | 0.00000 | 0.00000 | 0.0736 | 0.0587 | 0.0667 | 0.1720 |
|     | 5     | 0.01874 | 0.00037 | 0.00034 | 0.00038 | 0.04067 | 0.00078 | 0.00071 | 0.00080 | 0.00161 | 0.00000 | 0.00000 | 0.00000 | 0.0633 | 0.0389 | 0.0471 | 0.1178 |
|     | 10    | 0.03348 | 0.00058 | 0.00051 | 0.00057 | 0.07302 | 0.00123 | 0.00109 | 0.00122 | 0.00321 | 4.0e-6  | 3.0e-6  | 4.0e-6  | 0.0575 | 0.0251 | 0.0297 | 0.0723 |
| MC  | 0     |     | 0.02721 | 0.02487 | 0.01486 |     | 0.05648 | 0.05128 | 0.03047 |     | 0.00218 | 0.00171 | 0.00076 |     | 0.0144 | 0.0125 | 0.0333 |
|     | 5     |     | 0.02702 | 0.02522 | 0.01760 |     | 0.05687 | 0.05314 | 0.03713 |     | 0.00201 | 0.00186 | 0.00098 |     | 0.0103 | 0.0126 | 0.0346 |
|     | 10    |     | 0.02825 | 0.02504 | 0.01600 |     | 0.06002 | 0.05318 | 0.03409 |     | 0.00216 | 0.00177 | 0.00091 |     | 0.0094 | 0.0090 | 0.0295 |
| CPCS360 | | | | | | | | | | | | | | | | | |
| 1   | 10    | 0.26421 | 0.14222 | 0.13907 | 0.14334 | 7.78167 | 2119.20 | 2132.78 | 2133.84 | 0.17974 | 0.09297 | 0.09151 | 0.09255 | 0.7172 | 0.5486 | 0.5282 | 0.4593 |
|     | 20    | 0.26326 | 0.12867 | 0.12937 | 0.13665 | 370.444 | 28720.38 | 30704.93 | 31689.59 | 0.17845 | 0.08212 | 0.08269 | 0.08568 | 0.6794 | 0.5547 | 0.5250 | 0.4578 |
| 10  | 10    | 0.01772 | 0.00694 | 0.00121 | 0.00258 | 1.06933 | 6.07399 | 0.01005 | 0.04330 | 0.017718 | 0.00203 | 0.00019 | 0.00116 | 7.2205 | 4.7781 | 4.5191 | 3.7906 |
|     | 20    | 0.02413 | 0.00466 | 0.00115 | 0.00138 | 62.99310 | 26.04308 | 0.00886 | 0.01353 | 0.02027 | 0.00118 | 0.00015 | 0.00036 | 7.0830 | 4.8705 | 4.6468 | 3.8392 |
| 20  | 10    | 0.01772 | 0.00003 | 3.0e-6  | 3.0e-6  | 1.06933 | 0.00044 | 8.0e-6  | 7.0e-6  | 0.01771 | 5.0e-6  | 0.0     | 0.0     | 14.4379 | 9.5783 | 9.0770 | 7.6017 |
|     | 20    | 0.02413 | 0.00001 | 9.0e-6  | 9.0e-6  | 62.9931 | 0.00014 | 0.00013 | 0.00004 | 0.02027 | 0.0     | 0.0     | 0.0     | 13.6064 | 9.4582 | 9.0423 | 7.4453 |
| MC  | 10    |     | 0.03389 | 0.01984 | 0.01402 |     | 0.65600 | 0.20023 | 0.11990 |     | 0.01299 | 0.00590 | 0.00390 |     | 2.8077 | 2.7112 | 2.5188 |
|     | 20    |     | 0.02715 | 0.01543 | 0.00957 |     | 0.81401 | 0.17345 | 0.09113 |     | 0.01007 | 0.00444 | 0.00234 |     | 2.8532 | 2.7032 | 2.5297 |

based Patient Case Simulation system, known to be hard for belief updating) and on coding networks. On each type of networks, we ran Iterative Belief Propagation (IBP), MC(i) and IJGP(i), while giving IBP and IJGP(i) the same number of iterations.

We use the partitioning method described in Section 5 to construct a join-graph. To determine the order of message computation, we recursively pick an edge (u,v), such that node u has the fewest incoming messages missing.

For each network except coding, we compute the exact solution and compare the accuracy of algorithms using: 1. Absolute error - the absolute value of the difference between the approximate and the exact, averaged over all values, all variables and all problems. 2. Relative error - the absolute value of the difference between the approximate and the exact, divided by the exact, averaged over all values, all variables and all problems. 3. KL distance - $P_{exact}(X=a) \cdot log(P_{exact}(X=a)/P_{approximation}(X=a))$ averaged over all values, all variables and all problems. We also report the time taken by each algorithm. For coding networks we report Bit Error Rate (BER) computed as follows: for each approximate algorithm we pick the most likely value for each variable, take the number of disagreements with the exact input, divide by the total number of variables, and average over all the instances of the problem. We also report time.

The random networks were generated using parameters (N,K,C,P), where N is the number of variables, K is their domain size, C is the number of conditional probability tables (CPTs) and P is the number of parents in each CPT. Parents in each CPT are picked randomly and each CPT is filled randomly. In grid networks, N is a square number and each CPT is filled randomly. In each problem class, we also tested different numbers of evidence variables. The coding networks are from the class of linear block codes, where $\sigma$ is the channel noise level. Note that we are limited to relatively small and sparse problem instances since our evaluation measured are based on comparing against exact figures.

**Random network** results with networks of N=50, K=2, C=45 and P=3 are given in Table 1. For IJGP(i) and MC(i) we report 3 different values of i-bound: 2,

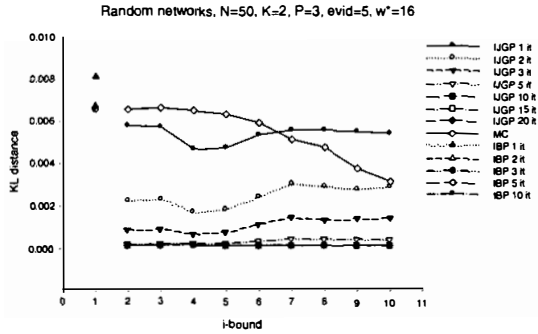

Figure 8: KL distance vs. i-bound

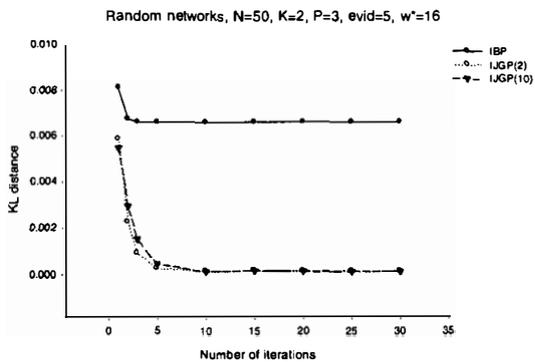

Figure 9: KL distance vs. iteration

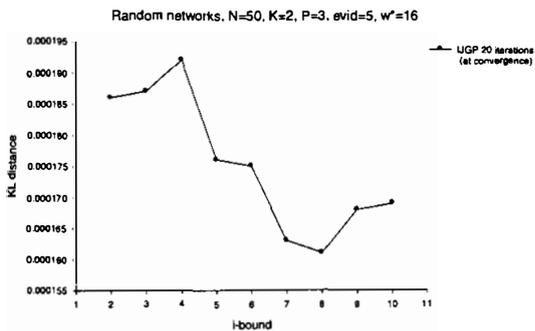

Figure 10: KL distance at convergence

5, 8; for IBP and IJGP(i) we report 3 different values of number of iterations: 1, 5, 10; for all algorithms we report 3 different values of number of evidence: 0, 5, 10. We notice that IJGP(i) is always better than IBP (except when i=2 and number of iterations is 1), sometimes as much as an order of magnitude, in terms of absolute and relative error and KL distance. IBP rarely changes after 5 iterations, whereas IJGP(i) solution can be improved up to 15-20 iterations. As we predicted, IJGP(i) is about equal to MC(i) in terms of accuracy for one iteration. But IJGP(i) improves as the number of iterations increases, and is eventually better than MC(i) by as much as an order of magnitude.
It clearly takes more time when the i-bound is large. Figure 8 shows a comparison of all algorithms with different numbers of iterations, using the KL distance.

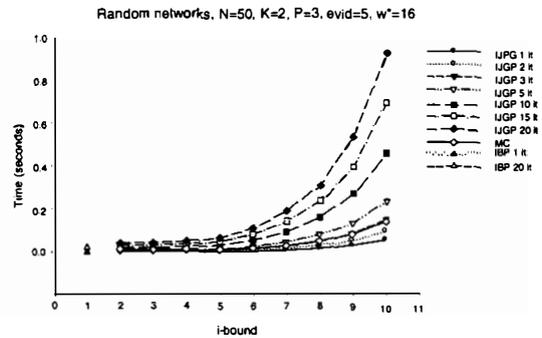

Figure 11: Time vs. i-bound

Because the network structure changes with different i-bounds, we do not see monotonic improvement of IJGP with i-bound for a given number of iterations (as is the case with MC). When IJGP converges it seems to yield constant error as a function of the i-bound (Figure 8), but on a higher resolution we notice a general trend of improvement with i-bound, as in Figure 10, demonstrating its anytime characteristic. We see that this feature is consistent throughout our experiments. Figure 9 shows how IJGP converges with iteration to smaller KL distance than IBP. As expected, the time taken by IJGP (and MC) varies exponentially with the i-bound (see Figure 11).

**Grid network** results with networks of N=81, K=2,

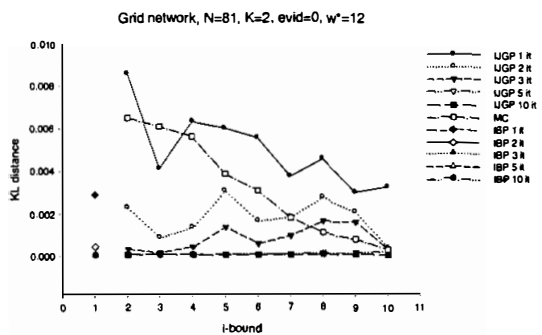

Figure 12: KL distance vs. i-bound (grid81, evid=0)

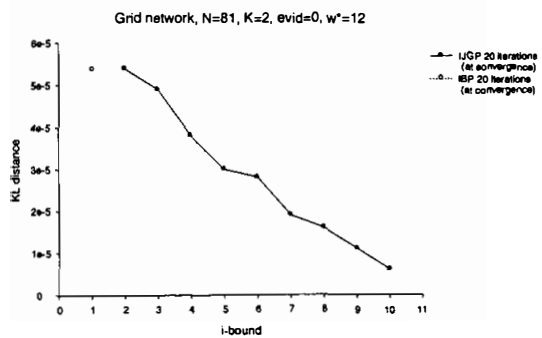

Figure 13: KL distance at convergence (grid81, evid=0)

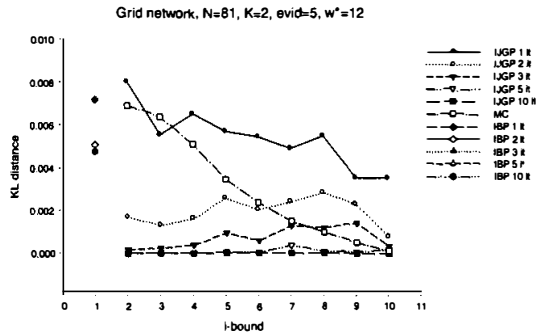

Figure 14: KL distance vs. i-bound (grid81, evid=5)

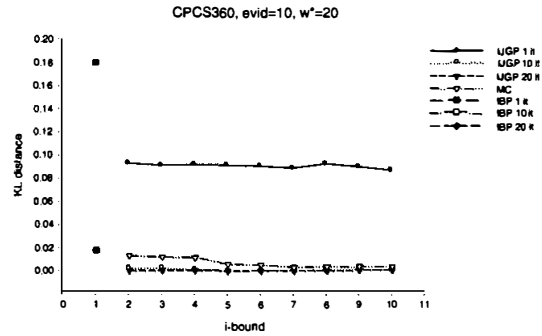

Figure 16: KL distance (CPCS360, evid=10)

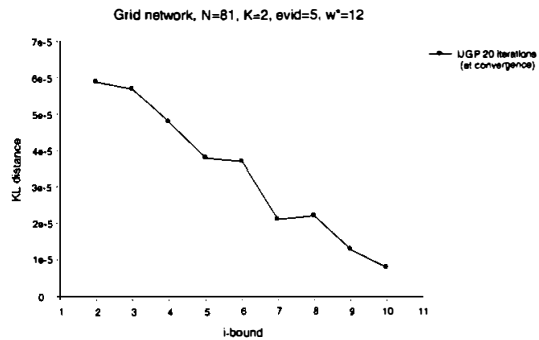

Figure 15: KL distance at convergence (grid81, evid=5)

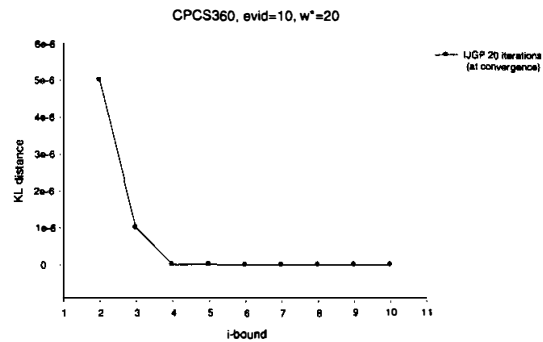

Figure 17: KL distance (CPCS360, evid=10)

## 7 DISCUSSION AND CONCLUSION

100 instances are very similar to those of random networks. They are reported in Figures 12-15 (table results are omitted for space reasons), where we can see the impact of having evidence (0 and 5 evidence variables) on the algorithms. IJGP at convergence gives the best performance in both cases, while IBP's performance deteriorates with more evidence and is surpassed by MC with i-bound 5 or larger.

**CPCS network** results with CPCS54 and CPCS360 are given in Table 2, and are even more pronounced than those of random and grid networks. When evidence is added, IJGP(i) is more accurate than MC(i), which is more accurate than IBP, as can be seen in Figure 16. Notice that the time is not changing much as the i-bound increases for CPCS360 networks. One reason may be due to the existence of functions with large scopes, which force large clusters even when i=2.

**Coding network** results are given in Table 3. We tested a large network of 400 variables, with tree-width $w^*=43$, with IJGP and IBP set to run 30 iterations. IBP is known to be very accurate for this class of problems and it is indeed better than MC. It is remarkable however that IJGP converges to smaller BER than IBP even for small values of the i-bound. Both the coding network and CPCS360 show the scalability of IJGP for large size problems. Notice that here the anytime behavior of IJGP is not clear.

The paper presents an iterative anytime approximation algorithm called Iterative Join-Graph Propagation (IJGP(i)), that applies the message passing algorithm of join-tree clustering to join-graphs rather than join-trees, iteratively. The algorithm borrows the iterative feature from Iterative Belief Propagation (IBP) on one hand and is inspired by the anytime virtues of mini-clustering MC(i) on the other. We show that the success of IJGP is facilitated by extending the notion of join-graphs to minimal arc-labeled join-graphs, and provide a structuring algorithm that generates minimal arc-labeled join-graphs of bounded size.

Our empirical results are extremely encouraging. We experimented with randomly generated networks, with grid-like networks, with medical diagnosis CPCS networks and with coding networks. We showed that IJGP is almost always superior to both IBP and MC(i) and is sometimes more accurate by an order of several magnitudes. One should note that IBP cannot be improved with more time, while MC(i) requires a large i-bound for many hard and large networks to achieve reasonable accuracy. There is no question that the iterative application of IJGP is instrumental to its success. In fact, IJGP(2) in isolation appears to be the most cost effective variant.

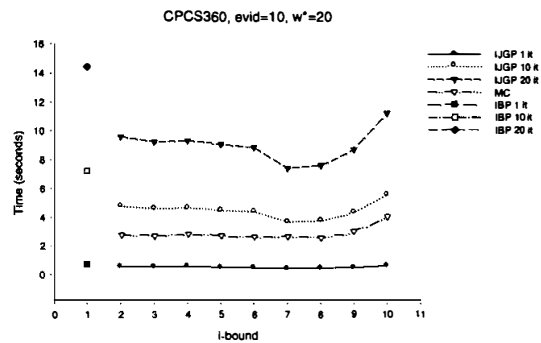

Figure 18: Time (CPCS360, evid=10)

Table 3: Coding networks: N=400, P=4, 500 instances, 30 iterations, w*=43

| $\sigma$ | | \multicolumn{5}{c|}{Bit Error Rate i-bound} | IBP |
| --- | --- | --- | --- | --- | --- | --- | --- |
| | | 2 | 4 | 6 | 8 | 10 | |
| 0.22 | IJGP | 0.00005 | 0.00005 | 0.00005 | 0.00005 | 0.00005 | 0.00005 |
| | MC | 0.00501 | 0.00800 | 0.00586 | 0.00462 | 0.00392 | |
| 0.28 | IJGP | 0.00062 | 0.00062 | 0.00062 | 0.00062 | 0.00062 | 0.00064 |
| | MC | 0.02170 | 0.02968 | 0.02492 | 0.02048 | 0.01840 | |
| 0.32 | IJGP | 0.00238 | 0.00238 | 0.00238 | 0.00238 | 0.00238 | 0.00242 |
| | MC | 0.04018 | 0.05004 | 0.04480 | 0.03878 | 0.03558 | |
| 0.40 | IJGP | 0.01202 | 0.01188 | 0.01194 | 0.01210 | 0.01192 | 0.01220 |
| | MC | 0.08726 | 0.09762 | 0.09272 | 0.08766 | 0.08334 | |
| 0.51 | IJGP | 0.07664 | 0.07498 | 0.07524 | 0.07578 | 0.07554 | 0.07816 |
| | MC | 0.15396 | 0.16048 | 0.15710 | 0.15452 | 0.15180 | |
| 0.65 | IJGP | 0.19070 | 0.19056 | 0.19016 | 0.19030 | 0.19056 | 0.19142 |
| | MC | 0.21890 | 0.22056 | 0.21928 | 0.21904 | 0.21830 | |
| | | \multicolumn{5}{c|}{Time} | |
| | IJGP | 0.36262 | 0.41695 | 0.86213 | 2.62307 | 9.23610 | 0.019752 |
| | MC | 0.25281 | 0.21816 | 0.31094 | 0.74851 | 2.33257 | |

One question that remains unanswered is why propagating the messages iteratively helps. Why is IJGP upon convergence, superior to IJGP with one iteration and is superior to MC(i)? One clue can be provided when considering deterministic constraint networks which can be viewed as "extreme probabilistic networks". It is known that constraint propagation algorithms, which are analogous to the messages sent by belief propagation, are guaranteed to converge and are guaranteed to improve with convergence. The propagation scheme presented here works like constraint propagation relative to the flat abstraction of $P$, (where all non-zero entries are normalized to a positive constant), and is guaranteed to be more accurate for that abstraction at least. Understanding the general case is still an open question.

## Acknowledgments

This work was supported in part by NSF grant IIS-0086529 and by MURI ONR award N00014-00-1-0617.

## References

[Arnborg1985] S. A. Arnborg. Efficient algorithms for combinatorial problems on graphs with bounded decomposability - a survey. *BIT*, 25:2–23, 1985.

[Cooper1990] G.F. Cooper. The computational complexity of probabistic inferences. *Artificial Intelligence*, pages 393–405, 1990.

[Dagum and Luby1993] P. Dagum and M. Luby. Approximating probabilistic inference in bayesian belief networks is NP-hard. In *National Conference on Artificial Intelligence (AAAI-93)*, 1993.

[Dechter and Rish1997] R. Dechter and I. Rish. A scheme for approximating probabilistic inference. In *Artificial Intelligence (UAI'97)*, pages 132–141, 1997.

[Dechter1996] R. Dechter. Bucket elimination: A unifying framework for probabilistic inference algorithms. In *Uncertainty in Artificial Intelligence (UAI'96)*, pages 211–219, 1996.

[Gottlob et al.1999] G. Gottlob, N. Leone, and F. Scarello. A comparison of structural CSP decomposition methods. *IJCAI-99*, 1999.

[Jensen et al.1990] F.V. Jensen, S.L Lauritzen, and K.G. Olesen. Bayesian updating in causal probabilistic networks by local computation. *Computational Statistics Quarterly*, 4:269–282, 1990.

[Larrosa et al.2001] J. Larrosa, K. Kask, and R. Dechter. Up and down mini-bucket: a scheme for approximating combinatorial optimization tasks. *Submitted*, 2001.

[Lauritzen and Spiegelhalter1988] S.L. Lauritzen and D.J. Spiegelhalter. Local computation with probabilities on graphical structures and their application to expert systems. *Journal of the Royal Statistical Society, Series B*, 50(2):157–224, 1988.

[Maier1983] D. Maier. The theory of relational databases. In *Computer Science Press, Rockville, MD*, 1983.

[Mateescu et al.2002] R. Mateescu, R. Dechter, and K. Kask. Tree approximation for belief updating. In *AAAI-02, to appear*, 2002.

[Pearl1988] J. Pearl. *Probabilistic Reasoning in Intelligent Systems*. Morgan Kaufmann, 1988.

[Roth1996] D. Roth. On the hardness of approximate reasoning. *AI*, 82(1-2):273–302, April 1996.

[Welling and Teh2001] M. Welling and Y.W. Teh. Belief optimization for binary networks: A stable alternative to loopy belief propagation. In *UAI'01*, pages 554–561, 2001.

[Yedidia et al.2001] J. S. Yedidia, W.T. Freeman, and Y. Weiss. Generalized belief propagation. In *Advances in Neural Information Processing Systems 13*, 2001.